\documentclass[anonymous, journal]{IEEEtran}

\usepackage{caption}
\usepackage{url}
\usepackage[hidelinks]{hyperref}
\hypersetup{
    colorlinks=true,
    linkcolor=blue,
    citecolor=blue,      
    urlcolor=blue,
}
\usepackage{amsmath}
\usepackage{tikz}
\usetikzlibrary{shapes.geometric, arrows, positioning}
\usepackage{graphicx}
\usepackage{algorithm}
\usepackage{algpseudocode}
\usepackage{url}

\usepackage{lscape}
\usepackage{booktabs}
\usepackage{rotating}
\usepackage{colortbl}
\usepackage[table,xcdraw]{xcolor}
\usepackage{graphicx}
\usepackage{algorithm}
\usepackage{algpseudocode}
\usepackage{stfloats}
\usepackage{enumitem}
\usepackage{multirow}
\usepackage{float}
\usepackage{subcaption}
\usepackage{comment}
\usepackage[table]{xcolor} 
\usepackage{orcidlink}
\usepackage[numbers,sort&compress]{natbib}
\def\BibTeX{{\rm B\kern-.05em{\sc i\kern-.025em b}\kern-.08em
    T\kern-.1667em\lower.7ex\hbox{E}\kern-.125emX}}
\usepackage{xcolor}

\begin{document}

\title{Domain-Informed Genetic Superposition Programming: A Case Study on SFRC Beams}

 \author{Mohammad~Sadegh~Khorshidi\orcidlink{0000-0001-6556-2926}% <-this % stops a space
 %\href{https://orcid.org/0000-0001-6556-2926}{{\aiOrcid}}% <-this % stops a space
 , Navid~Yazdanjue\orcidlink{0000-0001-9670-8422}%
 , Hassan~Gharoun\orcidlink{0000-0001-8298-7512}%
 , Mohammad~Reza~Nikoo\orcidlink{0000-0002-3740-4389}%
 , Fang~Chen\orcidlink{0000-0003-4971-8729}%
 , Amir~H.~Gandomi\orcidlink{0000-0002-2798-0104}
         % <-this % stops a space
 \thanks{Mohammad Sadegh Khorshidi, Navid Yazdanjue, Hassan Gharoun, Fang Chen, and Amir H. Gandomi are with the Faculty of Engineering \& Information Technology, University of Technology Sydney, Ultimo 2007, Australia. e-mails: msadegh.khorshidi.ak@gmail.com, navid.yazdanjue@gmail.com, hassan.gharoun@student.uts.edu.au, fang.chen@uts.edu.au, gandomi@uts.edu.au}% <-this % stops a space
 \thanks{Mohammad Reza Nikoo is with the Department of Civil and Architectural Engineering, Sultan Qaboos University, Muscat, Oman. (e-mail: m.reza@squ.edu.om}%
 \thanks{Amir H. Gandomi is also with the University Research and Innovation Center (EKIK), Obuda University, Budapest 1034, Hungary.}%
 \thanks{This work was supported by the Australian Government through the Australian Research Council under Project DE210101808.}
 \thanks{Corresponding author: Amir H. Gandomi}}

%\markboth{IEEE Transactions on Evolutionary Computation}%
\maketitle

\begin{abstract}
This study presents \textcolor{black}{domain-informed genetic superposition programming} (DIGSP), a symbolic regression framework tailored for engineering systems governed by separable physical mechanisms. DIGSP partitions the input space into domain-specific feature subsets and evolves independent \textcolor{black}{genetic programming (GP)} populations to model material-specific effects. Early evolution occurs in isolation, while ensemble fitness promotes inter-population cooperation. To enable symbolic superposition, an \textcolor{black}{adaptive hierarchical symbolic abstraction mechanism} (AHSAM) is triggered after stagnation across all populations. AHSAM performs \textcolor{black}{analysis of variance- (ANOVA) based} filtering to identify statistically significant individuals, compresses them into symbolic constructs, and injects them into all populations through a validation-guided pruning cycle. \textcolor{black}{The DIGSP is benchmarked} against a \textcolor{black}{baseline multi-gene genetic programming (BGP)} model using a dataset of steel fiber-reinforced concrete (SFRC) beams. Across 30 independent trials with 65\% training, 10\% validation, and 25\% testing splits, DIGSP consistently outperformed BGP in training and test \textcolor{black}{root mean squared error (RMSE)}. The Wilcoxon rank-sum test confirmed statistical significance ($p < 0.01$), and DIGSP showed tighter error distributions and fewer outliers. No significant difference was observed in validation RMSE due to limited sample size. These results demonstrate that domain-informed structural decomposition and symbolic abstraction improve convergence and generalization. DIGSP offers a principled and interpretable modeling strategy for systems where symbolic superposition aligns with the underlying physical structure.

\end{abstract}


\section{Introduction} \label{Section:Introduction}
\IEEEPARstart{T}{he} symbolic regression (SR) methods—particularly those based on genetic programming (GP)—have emerged as powerful tools for deriving interpretable models in engineering domains characterized by complex, nonlinear, and often non-parametric relationships \citep{angelis2023artificial, gandomi2012new, vaddireddy2020feature}. In structural engineering, predicting the shear strength of steel fiber-reinforced concrete (SFRC) beams exemplifies this complexity, as it involves highly nonlinear interactions between mechanical, geometrical, and material properties \citep{gandomi2011nonlinear, megahed2025symbolic}. Traditional empirical or semi-empirical models often rely on simplifications and pre-assumed functional forms \textcolor{black}{that limits} their generalizability across material variations and loading regimes \citep{aragon2025developing, batista2025embedding}. SR, by contrast, enables the automatic discovery of closed-form equations from data while preserving physical interpretability and dimensional consistency \citep{cohen2024physics, oh2023genetic}.

The advantages of symbolic modeling in civil and structural engineering are underscored by recent contributions employing physics-informed GP frameworks to uncover governing equations for stress-strain behavior and load-bearing capacities \citep{chadalawada2020hydrologically, carvalho2024physics, aragon2025developing}. Multi-gene GP structures and constrained optimization \citep{khorshidi2024integrating, khorshidi2024agent} have further enhanced modeling robustness and equation parsimony \citep{luan2023physics, khorshidi2023filter}. Yet, despite these advances, a key challenge persists: the heterogeneous nature of the input features—including fiber geometry, mix composition, and mechanical reinforcement—introduces redundant or conditionally independent dimensions that often impair model convergence, generalizability, or interpretability when processed as a monolithic input set \citep{khorshidi2025semantic, fleck2024population}.

In many civil engineering problems, such as SFRC beam modeling, the principle of superposition is fundamental—predictive responses can be approximated as additive or interacting contributions from distinct physical mechanisms. By leveraging domain information to separate input features into semantically coherent groups (e.g., fiber parameters, concrete characteristics, geometric dimensions), the learning algorithm \textcolor{black}{is enabled} to model such sub-mechanisms independently. This separation allows the system to evolve localized symbolic expressions that are structurally simpler and more interpretable. When appropriately combined using symbolic abstraction, these expressions exhibit superimposed behavior aligned with physical intuition, reducing both search complexity and model redundancy.

To realize this synergy, a new SR framework termed \textcolor{black}{domain-informed genetic superposition programming} (DIGSP), \textcolor{black}{is proposed} where domain-informed feature partitioning enables independent modeling of separable mechanisms, and symbolic abstraction allows their superposition into holistic expressions. DIGSP departs from conventional GP by structuring its population into physically meaningful subgroups that co-evolve in parallel while maintaining modularity. 

To further enhance convergence efficiency, DIGSP employs the \textcolor{black}{hierarchical symbolic abstraction mechanism} (AHSAM) \citep{khorshidi2025transformer}, which activates upon stagnation and identifies statistically significant symbolic expressions using \textcolor{black}{analysis of variance} (ANOVA). These expressions are abstracted into high-level features and injected into subsequent generations to guide symbolic search. The AHSAM is adapted from an earlier work on symbolic modeling of transformer embeddings for interpretable classification \citep{khorshidi2025transformer}, where abstraction enabled generalization in high-dimensional and sparsely labeled domains.

The \textcolor{black}{proposed} methodology contributes a principled solution to the dual challenge of representation and optimization in symbolic modeling of SFRC beams shear strength. Specifically:
\begin{itemize}
    \item The DIGSP framework \textcolor{black}{is introduced}, wherein input features are pre-partitioned based on engineering semantics (e.g., fiber properties, concrete mix, geometry), and evolved in isolated subpopulations.
    \item The AHSAM \textcolor{black}{is applied}, which activates upon stagnation to compress symbolic knowledge across subpopulations using statistically significant abstraction and reinjection.
    \item Symbolic superposition of evolved expressions \textcolor{black}{is enabled}, allowing DIGSP to integrate cross-domain information while preserving modularity and minimizing overfitting.
    \item Adaptive structural pruning \textcolor{black}{is integrated} to ensure that injected abstractions do not degrade model accuracy, thus facilitating efficient reuse of abstracted patterns.
    \item Demonstrate the capability of DIGSP to produce compact, generalizable, and interpretable symbolic models in a high-complexity engineering task.
\end{itemize}

This study aligns with a growing body of work advocating for physically informed symbolic modeling as a scalable alternative to black-box regression, particularly in high-stakes engineering contexts where interpretability, parsimony, and domain compatibility are essential \citep{faroughi2024physics, de2024numerical, reis2024benchmarking}. Our contributions establish a novel synthesis between domain-guided feature separation, symbolic superposition, and hierarchical abstraction, expanding the toolkit of interpretable machine learning for engineering systems.

The remainder of this paper is structured as follows. Section~\ref{Section:Related_Works} reviews related literature on SR, domain-informed modeling, and feature partitioning. \textcolor{black}{Section~\ref{sec:case_study} presents a case study on shear strength modeling of SFRC beams, describing the dataset, domain-informed variable partitioning, and structural motivations.} Section~\ref{Section:Methodology} details the DIGSP framework, including its domain-informed multi-population structure and the AHSAM. Section~\ref{Section:Results_Discussion} presents a comparative evaluation of DIGSP against a \textcolor{black}{baseline genetic programming} (BGP) model using the SFRC beam dataset. Finally, Section~\ref{Section:Conclusion} concludes the paper with key findings and directions for future research.

\section{Related Works} \label{Section:Related_Works}
The use of GP and SR in engineering modeling has gained prominence as a viable alternative to black-box models due to its ability to produce human-interpretable expressions that align with domain physics. Early applications in structural engineering include Gandomi and Alavi's linear GP for modeling the shear strength of SFRC beams, achieving promising results in replicating nonlinear material behavior from experimental datasets \citep{gandomi2011nonlinear}. This work was later extended via a multi-gene GP approach that allows modular nonlinear substructures to be linearly combined with optimized weights, enabling improved accuracy and flexibility while \textcolor{black}{the complexity of the model is controlled} \citep{gandomi2012new}. Recent reviews confirm SR's advantages in transparency, physical consistency, and adaptability for nonlinear material systems, particularly under uncertainty \citep{angelis2023artificial, oh2023genetic}.

The evolution of SR methodologies has also embraced techniques that address overfitting, bloat control, and dynamic feature construction. Bomarito et al. proposed a Bayesian model selection mechanism to prune bloated SR structures and prevent overfitting in material modeling \citep{bomarito2022bayesian}. Similarly, modular GP frameworks, such as those proposed by Zhang et al., construct multi-tree individuals with separate submodules for feature interaction, which improves generalization while preserving transparency \citep{zhang2023modular}. More recently, adaptive population dynamics have been introduced to dynamically evolve model complexity in response to data characteristics during SR \citep{fleck2024population}. These developments align with efforts to capture physics-constrained relationships in materials science, such as governing stress-strain behavior or post-peak response in SFRC beams.

In parallel, feature partitioning and ensemble learning have been instrumental in reducing model variance, improving interpretability, and decreasing training cost, particularly in high-dimensional regression tasks. The concept of multi-view ensemble learning has provided a principled basis for splitting input features into conditionally independent subsets, each assigned to a base learner, and aggregating their outputs \citep{khorshidi2025semantic}. Kumar and Yadav employed a minimum spanning tree clustering algorithm to identify feature partitions based on pairwise correlations, thus constructing decorrelated views that enhance ensemble diversity and reduce overfitting \citep{kumar2024minimum}. Li et al. integrated domain ontologies into attribute partitioning to guide ensemble learning for cognitive modeling, showcasing the relevance of domain knowledge in feature division \citep{li2022novel}. Such methods allow controlled redundancy and support modular learning frameworks \textcolor{black}{that are capable of} leveraging complementarity between feature groups.

The intersection of physics-informed machine learning and SR has emerged as a dominant theme for modeling physical systems with limited or noisy data. Cohen et al. formulated an SR framework guided by physics-based residuals to discover partial differential equations from scarce measurements \citep{cohen2024physics}. Similarly, Aragon et al. developed a strength model of engineering materials by encoding known constitutive relationships into the GP formulation, enabling extrapolation beyond training ranges and enforcing dimensional consistency \citep{aragon2025developing}. The incorporation of physics via constraints or penalization schemes is seen as vital in data-scarce domains such as SFRC shear modeling, where empirical data may be sparse or biased toward specific reinforcement configurations.

The role of symbolic abstraction mechanisms has gained attention in recent years to systematically integrate domain knowledge into model evolution. Megahed proposed a code-constrained SR (C-SR) approach that incorporates engineering design codes (e.g., EC4, AISC360) into the SR pipeline via constrained expression trees \citep{megahed2025symbolic}. Similarly, Batista embedded domain-relevant feature combinations suggested by GPT-4o into the feature space prior to GP, enabling semantically enriched representations \citep{batista2025embedding}. These works demonstrate that domain knowledge, when systematically injected into the symbolic modeling process, can guide the search toward physically consistent and interpretable models.

\textcolor{black}{This} study builds on these developments by applying AHSAM, which formalizes symbolic feature abstraction at stagnation points. Unlike prior approaches that embed constraints from the outset, AHSAM hierarchically extracts statistically significant symbolic expressions (e.g., $z_1, z_2, \dots, z_n$) from the best-performing individuals across partitioned populations. These expressions are reused in future generations to compress the search space and introduce symbolic gene exchange across otherwise isolated populations. AHSAM is adopted from our earlier work in transformer-based symbolic modeling for classification tasks \citep{khorshidi2025transformer}, where abstraction improved generalization from sparse, high-dimensional data. While conceptually related to modular GP and domain-guided abstraction, AHSAM differentiates itself by leveraging population-wise statistical tests (e.g., ANOVA $p$-values) as gating mechanisms for abstraction, thus enabling adaptive behavior conditioned on the convergence dynamics of each population.

In engineering applications, especially where model interpretability and physical soundness are critical (e.g., SFRC shear strength prediction), AHSAM provides a means to bridge domain-driven symbolic discovery with ensemble-based learning. This strategy is particularly powerful when inputs are partitioned a priori based on material categories (e.g., concrete mix, steel type, fiber geometry), as demonstrated in Li et al. \citep{li2024can}, who showed that encoding domain separability in the learning structure yields superior generalization.

Furthermore, \textcolor{black}{the proposed} methodology aligns with broader trends in physics-informed machine learning where symbolic models and deep networks increasingly leverage domain constraints \citep{pateras2023taxonomic, faroughi2024physics}. While neural models such as \textcolor{black}{physics-informed neural networks} (PINNs) have dominated recent discussions \citep{cuomo2022scientific, marian2023physics}, the SR alternative offers a complementary path prioritizing interpretability, analytic tractability, and extrapolation capacity—especially in domains like civil engineering where code compliance and model transparency are non-negotiable.

In this context, the proposed DIGSP provides a novel SR paradigm that unifies domain-informed variable separation and superposition principle under a modular evolutionary architecture. By abstracting and recombining meaningful symbolic units through AHSAM, DIGSP can achieve interpretable and high-performing models in engineering tasks characterized by complex functional dependencies and structural heterogeneity.

\section{Case Study: Shear Strength Modeling of SFRC Beams} \label{sec:case_study}

The development of DIGSP is motivated by a fundamental modeling challenge in civil engineering: capturing the shear behavior of steel fiber-reinforced concrete (SFRC) beams through modular yet interpretable representations. Shear strength in SFRC systems arises from a complex interplay between fiber bridging, concrete matrix cohesion, and internal force redistribution through reinforcement. These mechanisms operate at different structural scales and are governed by distinct physical laws. Their combined effect, however, exhibits a characteristic structure, where each component contributes either additively or in an interacting form to the beam’s ultimate shear resistance. This formulation is emblematic of the superposition principle, which underpins many engineering analyses by allowing the system-level response to be decomposed into—and recomposed from—independent subeffects.

Modeling such systems requires a framework that does not merely fit nonlinear data but aligns with the physical reasoning engineers apply in practice. DIGSP is designed specifically for domains like this, where structural heterogeneity and symbolic interpretability are equally critical. To evaluate its performance, a benchmark dataset of SFRC beams compiled by Gandomi et al.~\cite{gandomi2011nonlinear} \textcolor{black}{is used to aseess the efficiency of DIGSP}. Table \ref{tab:variables} provides an overview of the modeled variables and their interpretations. The dataset contains 213 experimental instances of beams without stirrups, each described by a set of input variables reflecting geometry, material composition, reinforcement, and fiber characteristics, alongside the measured ultimate shear strength $V_u$ (kN).

\begin{table*}[ht]
\centering
\caption{Variables used for modeling shear strength of SFRC beams}
\label{tab:variables}
\begin{tabular}{llll}
\toprule
\textbf{Variable} & \textbf{Symbol} & \textbf{Unit} & \textbf{Description} \\
\midrule
Beam width & $a$ & mm & \textcolor{black}{Width of cross section} \\
Effective depth & $d$ & mm & \textcolor{black}{Distance from compression face to centroid of tensile steel} \\
Compressive strength of concrete & $f_c'$ & MPa & \textcolor{black}{Strength of concrete matrix} \\
Shear span to depth ratio & $a/d$ & – & \textcolor{black}{Structural geometry ratio influencing failure mode} \\
Longitudinal reinforcement ratio & $\rho$ & – & \textcolor{black}{Steel reinforcement as a fraction of cross-sectional area} \\
Maximum aggregate size & $d_f$ & mm & \textcolor{black}{Aggregate granularity influencing bond and crack path} \\
Fiber volume fraction & $V_f$ & \% & \textcolor{black}{Percentage of fibers in the mix} \\
Aspect ratio of fibers & $l_f/d_f$ & – & \textcolor{black}{Ratio of fiber length to diameter} \\
Tensile strength of fiber & $f_f$ & MPa & \textcolor{black}{Strength of steel fibers} \\
\bottomrule
\end{tabular}
\end{table*}

\subsection{Dataset Summary and Diagnostics}\label{sec:dataset_summary}
\begin{figure*}[t]
\centering
\captionsetup[subfigure]{justification=centering,labelformat=parens,aboveskip=2pt,belowskip=0pt}
% -------- Row 1: SST2G (3) + 20NG (2)
\begin{subfigure}[t]{0.19\textwidth}\centering
\includegraphics[width=\linewidth]{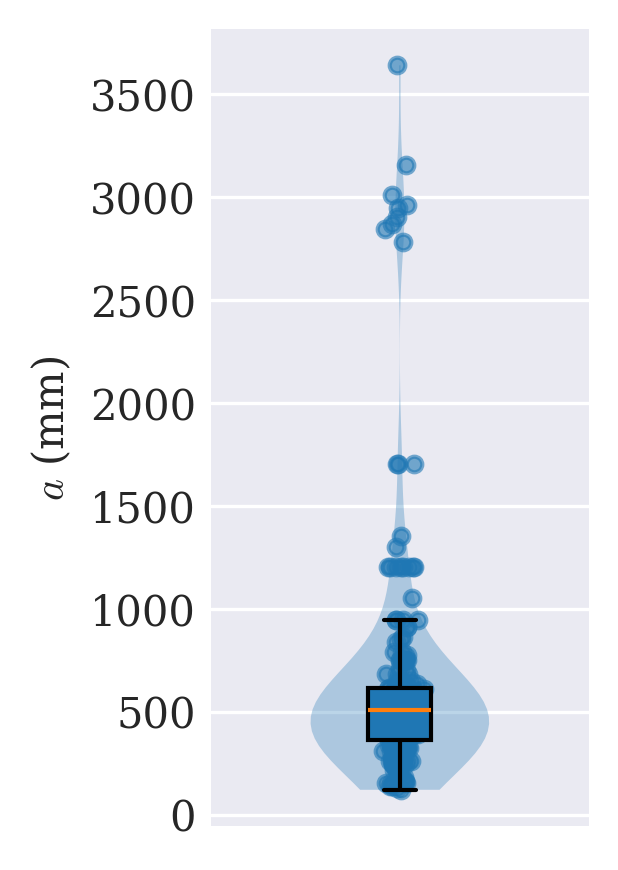}
\caption{$a$ (mm)}\label{subfig:a}
\end{subfigure}\hfill
\begin{subfigure}[t]{0.19\textwidth}\centering
\includegraphics[width=\linewidth]{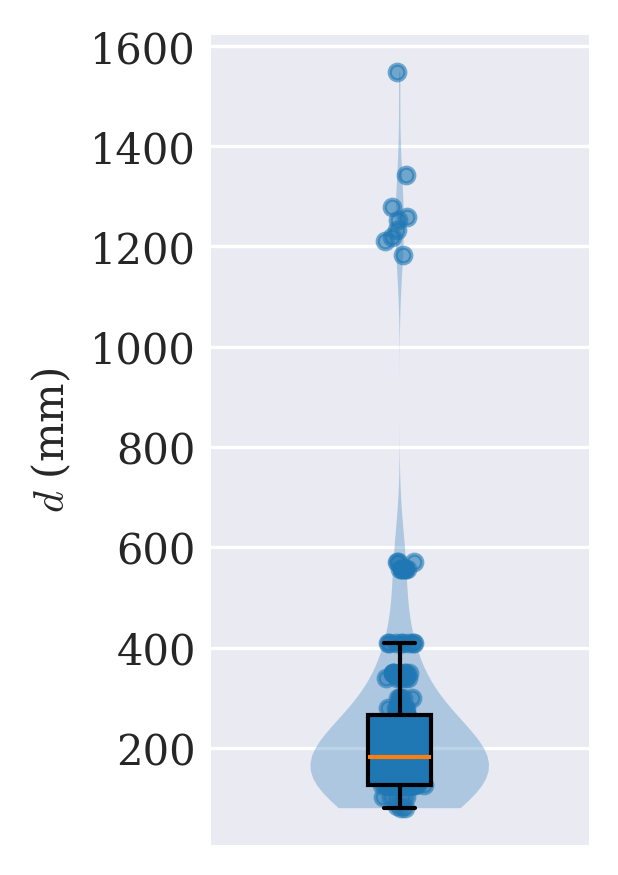}
\caption{$d$ (mm)}\label{subfig:d}
\end{subfigure}\hfill
\begin{subfigure}[t]{0.19\textwidth}\centering
\includegraphics[width=\linewidth]{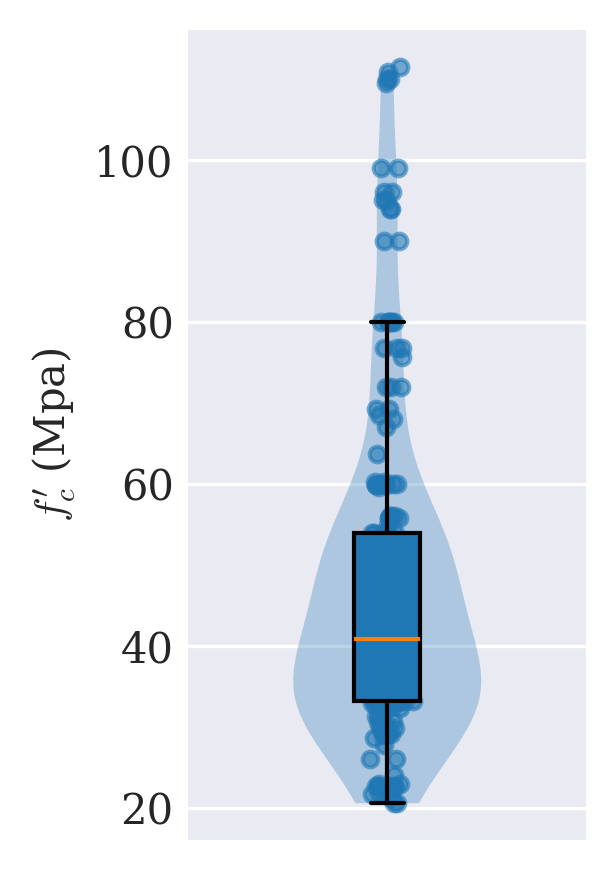}
\caption{$f_c'$ (MPa)}\label{subfig:fc}
\end{subfigure}\hfill
\begin{subfigure}[t]{0.19\textwidth}\centering
\includegraphics[width=\linewidth]{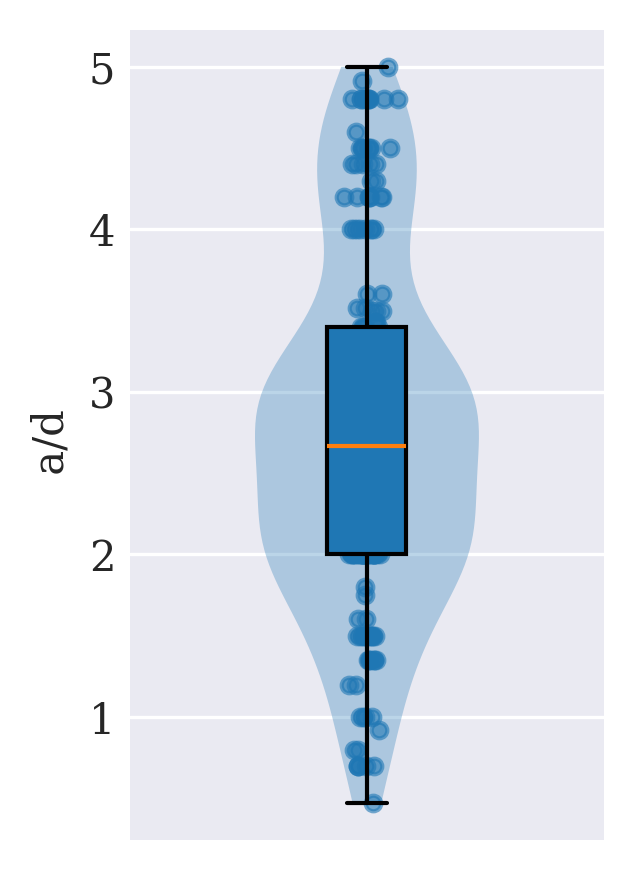}
\caption{$a/d$}\label{subfig:ad}
\end{subfigure}\hfill
\begin{subfigure}[t]{0.19\textwidth}\centering
\includegraphics[width=\linewidth]{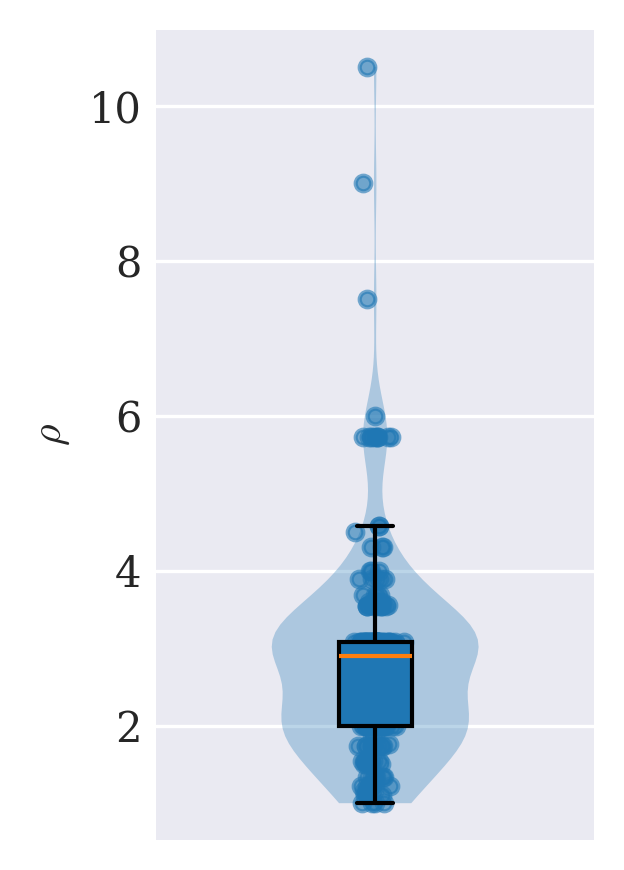}
\caption{$\rho$}\label{subfig:rho}
\end{subfigure}

\vspace{0.35em}

% -------- Row 2: 20NG (1) + MNIST (3) + CIFAR10 (1)
\begin{subfigure}[t]{0.19\textwidth}\centering
\includegraphics[width=\linewidth]{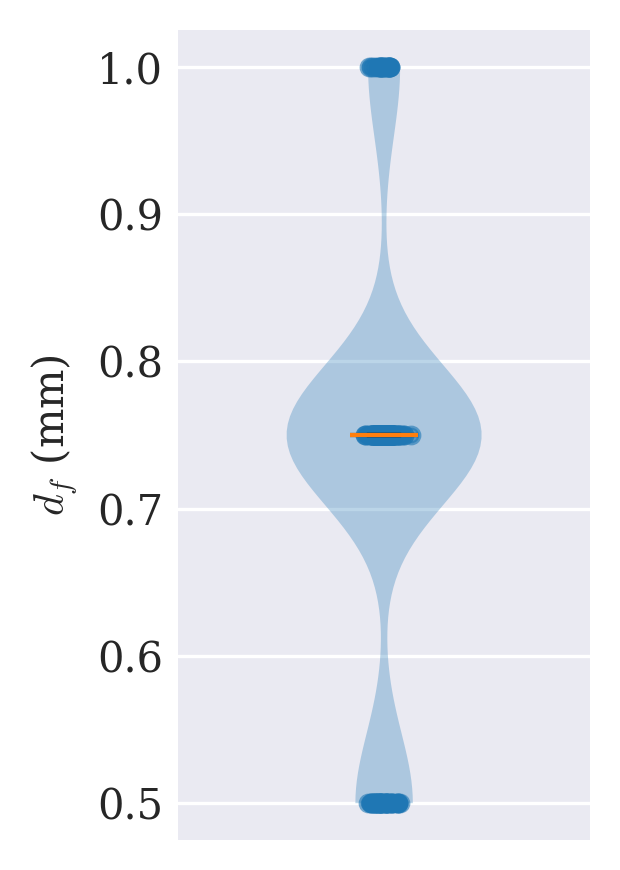}
\caption{$d_f$ (mm)}\label{subfig:df}
\end{subfigure}\hfill
\begin{subfigure}[t]{0.19\textwidth}\centering
\includegraphics[width=\linewidth]{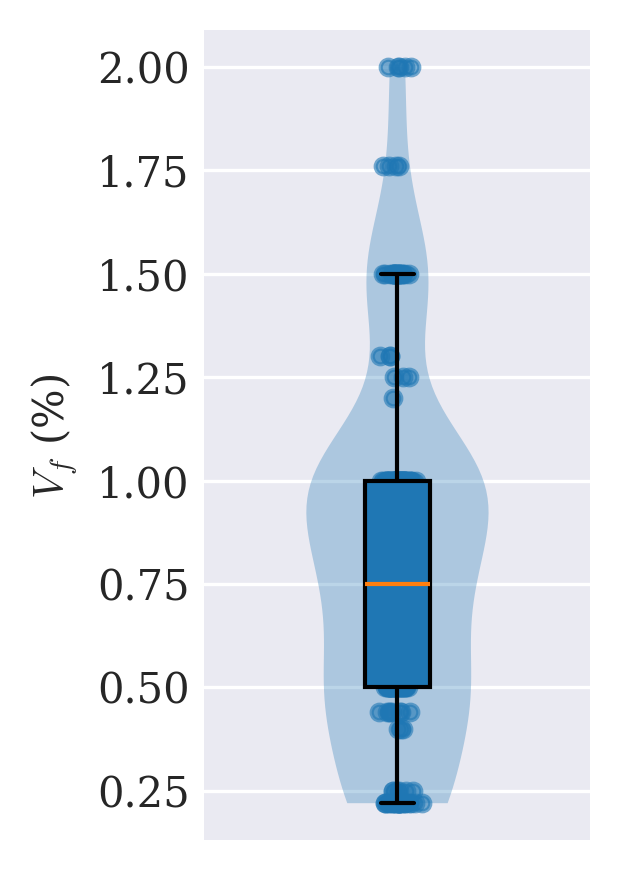}
\caption{$V_f$ (\%)}\label{subfig:vf}
\end{subfigure}\hfill
\begin{subfigure}[t]{0.19\textwidth}\centering
\includegraphics[width=\linewidth]{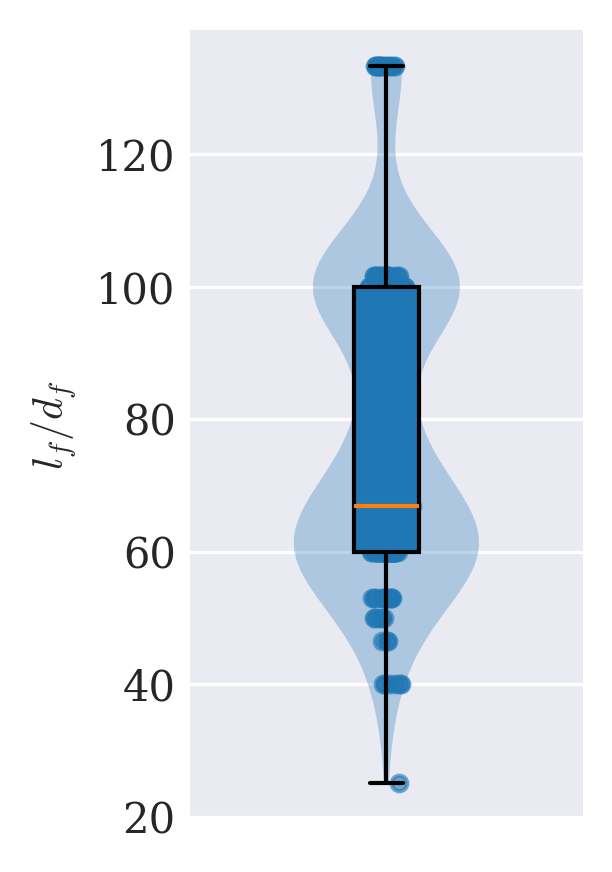}
\caption{$l_f/d_f$}\label{subfig:lfdf}
\end{subfigure}\hfill
\begin{subfigure}[t]{0.19\textwidth}\centering
\includegraphics[width=\linewidth]{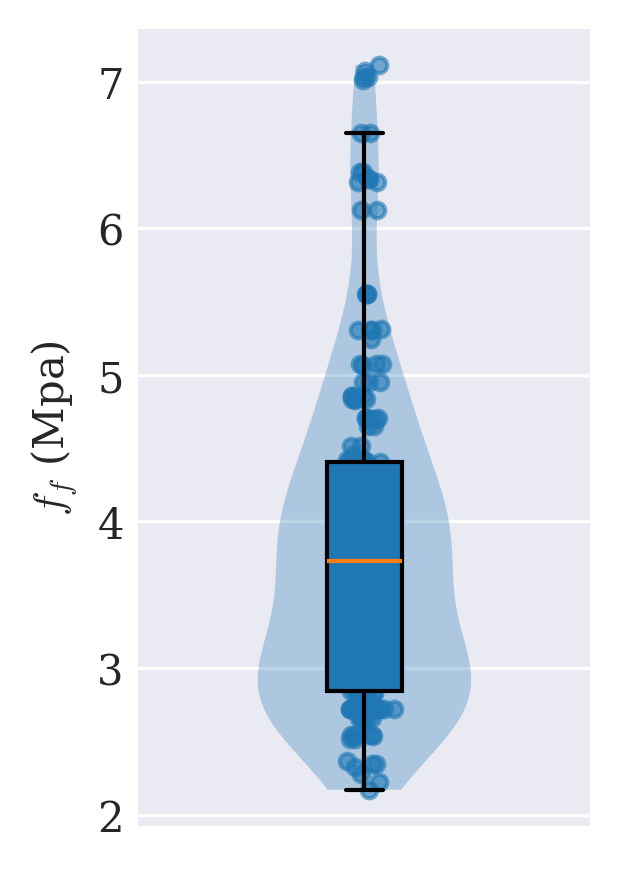}
\caption{$f_f$ (MPa)}\label{subfig:ff}
\end{subfigure}\hfill
\begin{subfigure}[t]{0.19\textwidth}\centering
\includegraphics[width=\linewidth]{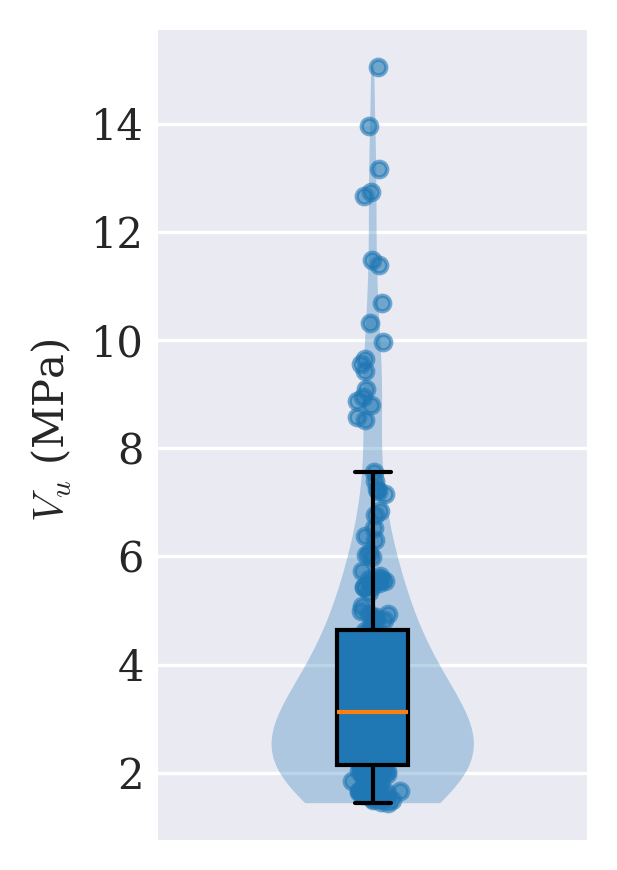}
\caption{$V_u$ (MPa)}\label{subfig:vu}
\end{subfigure}

\caption{\textcolor{black}{Distributions of the SFRC dataset variables (N=213 samples), (a) beam width, $a$ (mm), (b) effective depth, $d$ (mm), (c) compressive strength of concrete, $f_c'$ (MPa), (d) shear span to depth ratio, $a/d$, (e) longitudinal reinforcement ratio, $\rho$, (f) maximum aggregate size, $d_f$ (mm), (g) fiber volume fraction, $V_f$ (\%), (h) aspect ratio of fibers, $l_f/d_f$, (i) tensile strength of fiber, $f_f$ (MPa), and (j) the target variable, ultimate shearing strength $V_u$ (MPa).}}
\label{fig_data}
\end{figure*}

\textcolor{black}{The benchmark SFRC beam dataset (N=213 beams without stirrups) compiled by Gandomi et al.~\cite{gandomi2011nonlinear} is analyzed. The dataset is screened for missing values, duplicated records, and physically implausible entries (e.g., negative dimensions or strengths); no removals are required. Figure~\ref{fig_data} visualizes the empirical distributions of all input variables provided in Table \ref{tab:variables} and the target $V_u$ using violins with boxplot overlays and jittered points that provides a compact view of central tendency, dispersion, and rare observations. The ranges observed for geometry ($a$, $d$), material properties ($f_c'$, $f_f$, $d_f$), reinforcement ratio ($\rho$), fiber descriptors ($V_f$, $l_f/d_f$), and the response $V_u$ are consistent with reported experimental programs for SFRC beams and suggested adequate variability for model identification.}

\textcolor{black}{Table~\ref{tab:dataset_statistics} reports complementary numerical diagnostics (minimum, maximum, range, mean, standard deviation, median, skewness, and the percentage of Tukey outliers). Most variables exhibited moderate right-skew (e.g., $a$, $d$, $f_c'$, $\rho$, and $V_u$), while $d_f$ showed a narrow spread with a larger outlier rate that reflects its constrained measurement scale. Outlier rates are generally low (typically $<\!11\%$), and all units and magnitudes are physically plausible. These distributional characteristics motivated the domain-informed partitioning used in DIGSP (fiber$+$geometry, concrete$+$geometry, steel$+$geometry) and informed expectations of monotonic effects during symbolic search, while ensuring that evaluation of model performance is grounded in the empirical variability of the inputs.}

\begin{table*}[t]
\centering
\caption{\textcolor{black}{Summary diagnostics for the SFRC dataset (N=213) including, minimum, maximum, range, mean, standard deviation, median, skewness, and the percentage of outliers (Tukey’s 1.5$\times$\,IQR).}}
\label{tab:dataset_statistics}
\small
{%
\color{black}%
  \arrayrulecolor{black}% comment this line if you want black rules
\begin{tabular}{lccccccccc}
\toprule
\textbf{Variable} & \textbf{Unit} & \textbf{Minimum} & \textbf{Maximum} 
& \textbf{Range} & \textbf{Mean} & \textbf{Standard Deviation} & \textbf{Median} & \textbf{Skewness} & \textbf{\% Outlier}\\
\midrule
$b$ & mm & 120  & 3637.87  &  3517.87 & 633.14 & 574.2 & 508 & 3.202 & 10.8  \\
$d$ & mm   & 80  &  1548.03 & 1468.03  & 251.65 & 242.31 & 182 & 3.429 & 8.45  \\
$f_c'$ & MPa &  20.6 & 111.5  & 90.9  & 47.22 & 20.1 & 40.85 & 1.352 & 7.04  \\
$a/d$ & –   & 0.47  &  5 &  4.53 & 2.74 & 1.07 & 2.67 & 0.257 & 0  \\
$\rho$ & –  &  1 &  10.5 &  9.5 & 2.81 & 1.31 & 2.9 & 2.031 & 3.1  \\
$d_f$ & mm  &  0.5 &  1 &  0.5 & 0.73 & 0.14 & 0.75 & -0.038 & 31.46  \\
$V_f$ & \%  &  0.22 &  2 &  1.78 & 0.83 & 0.44 & 0.75 & 0.592 & 4.23  \\
$l_f/d_f$ & – &  25 &  133.33 & 108.33  & 79.43 & 24.48 & 66.84 & 0.504 & 0  \\
$f_f$ & MPa &  2.17 &  7.11 & 4.94  & 3.8 & 1.12 & 3.73 & 1.041 & 2.35  \\
$V_u$ & kN  &  1.44 &  15.05 &  13.61 & 3.93 & 2.6 & 3.13 & 1.919 & 8.92  \\
\bottomrule
\end{tabular}
}%
\end{table*}

\subsection{Domain-Informed Feature Partitioning and Structural Motivation}\label{sec:feature_partition}

To align symbolic modeling with physical mechanisms, the input variables are decomposed into three semantically distinct groups: (i) \textit{fiber and geometry}, comprising $V_f$, $l_f/d_f$, $a/d$, $a$, and $d$, which encode how fibers enhance shear performance through post-crack bridging and dimensional confinement; (ii) \textit{concrete and geometry}, comprising $f_c'$, $d_f$, $a/d$, and $d$, which control matrix shear capacity and crack path development; and (iii) \textit{steel and geometry}, comprising $\rho$, $a$, $d$, and $a/d$, which reflect how reinforcement and geometric ratios influence internal force paths and failure modes.

These groups are each assigned to a separate population in the GP model, such that structural mechanisms are independently modeled through population-specific symbolic expressions. Unlike traditional monolithic SR approaches, this partitioning allowed DIGSP to operate under a modular encoding regime that reflects the true decomposition of physical effects observed in beam behavior. Table~\ref{tab:variables} provides an overview of the modeled variables and their interpretations.

Each population is initialized with expressions formed exclusively from its assigned group, and no inter-population gene exchange is permitted during early evolution. This structure enforces the separation of causal domains, promoting the discovery of symbolic forms that capture the underlying physics without entangling independent effects. Once populations reached convergence stagnation, AHSAM is activated to extract and abstract the most statistically significant symbolic expressions. These abstractions are then injected back into the terminal sets of all populations, allowing the previously isolated mechanisms to be superimposed within the evolving symbolic forms.

By constructing symbolic models in this hierarchical and modular way, DIGSP operationalizes the superposition principle as a physical assumption and as a guiding logic for the search strategy. The resulting expressions reflect how fibers, matrix, and steel each contribute to shear strength in a compositional structure interpretable to engineering audiences. The framework is not bound to the SFRC domain—it is generalizable to any system in which physical separability and symbolic recomposition are present. The SFRC beam case study thus serves both as a canonical example and as a validation target for DIGSP’s symbolic architecture.

\section{Methodology} \label{Section:Methodology}

\begin{figure*}[!htbp]
  \centering
 \includegraphics[width=\textwidth]{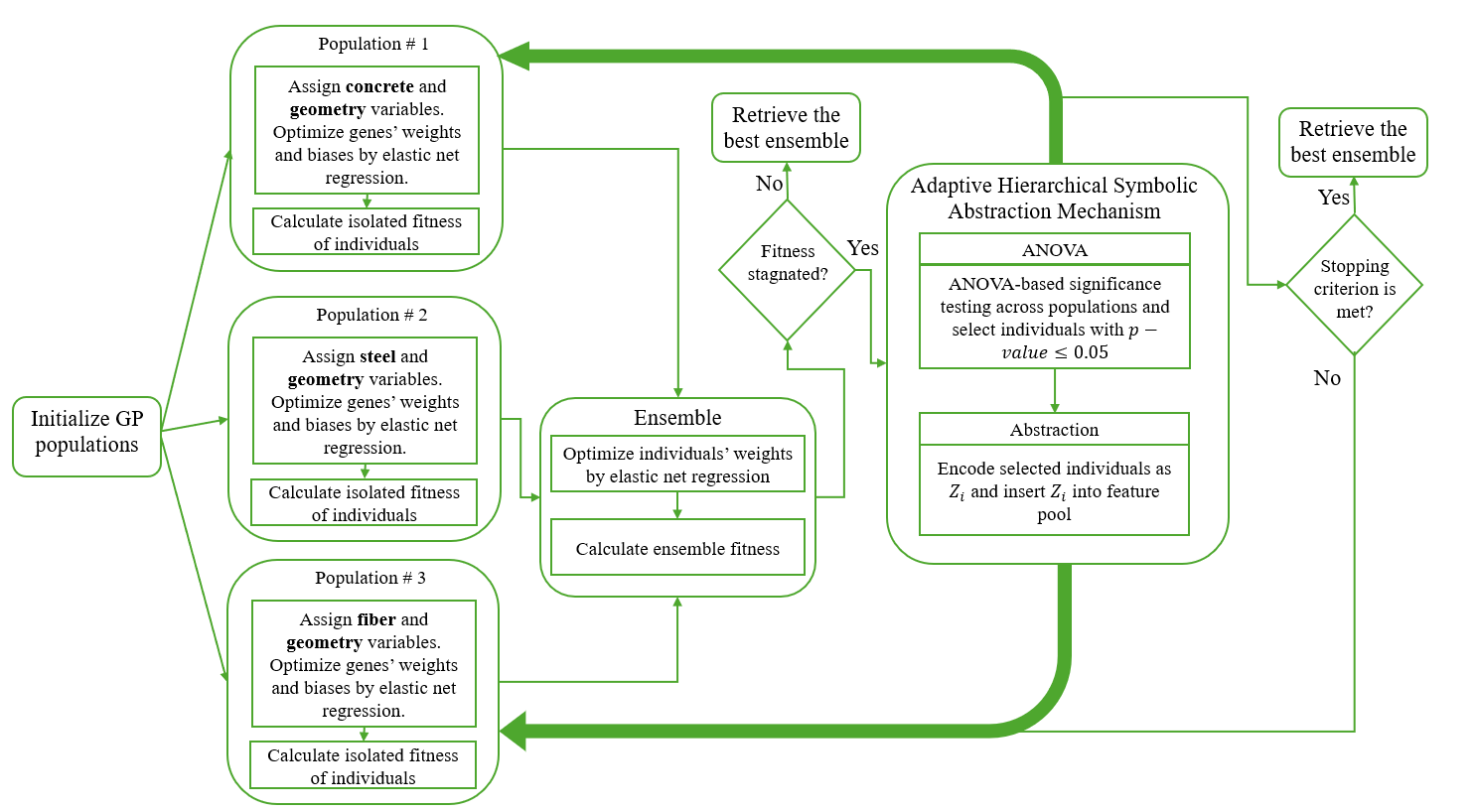}%
  \caption{Flowchart of the proposed DIGSP framework incorporating the \textcolor{blue}{AHSAM}.}
  \label{fig:flowchart}
\end{figure*}

DIGSP is a SR framework designed to model systems whose structural behavior adheres to the superposition principle. It achieves this by encoding symbolic components from distinct physical mechanisms independently and superimposing them through a modular abstraction hierarchy. This section outlines how DIGSP is constructed to capture shear strength behavior in SFRC beams and how its components generalize to other physical systems exhibiting separable structural effects. The full pipeline is illustrated in Figure~\ref{fig:flowchart}. \textcolor{black}{The diagram outlines key stages including initialization, feature-partitioned population evolution, ensemble-guided selection, symbolic regression via ridge optimization, and the AHSAM-triggered abstraction phase following convergence stagnation.}

\subsection{Domain-Informed Feature Partitioning and Population Structure}

As stated \textcolor{black}{in section} \ref{sec:feature_partition}, the input space is decomposed into three physically grounded subsets: (i) concrete and geometry (i.e. $f_c'$ and $a/d$), (ii) steel and geometry (i.e. $\rho$ and $a/d$), and (iii) fiber and geometry (i.e. $V_b$ and $a/d$. These subsets correspond to separable mechanisms that influence SFRC beam behavior through material cohesion, reinforcement dynamics, and post-crack fiber bridging. Each subset is assigned to a dedicated population, such that expressions within a population are evolved from a constrained terminal set reflective of its physical role.

This decomposition allows GP to evolve symbolic forms that reflect subsystem-specific behavior in isolation. Unlike standard multi-view learning approaches, the goal here is not only to enhance diversity but to instantiate physically interpretable partitions that mirror the assumptions underlying engineering superposition. By preventing inter-population gene exchange in early evolution, DIGSP enforces modular symbolic development before later stages of abstraction and recomposition.

\subsection{Gene Structure and Fitness Optimization via Elastic Net Regression}

Each individual consists of a vector of symbolic genes, where each gene is an expression tree. To produce an overall prediction, gene outputs are linearly aggregated via \textcolor{black}{elastic net regression} (ENR), balancing sparsity and regularization to avoid redundant or unstable symbolic terms.

Let $g_j(x_i)$ denote the output of gene $j$ for sample $x_i$, and $y_i$ the observed target. The predicted output is given by:
\begin{equation}
    \hat{y}_i = \sum_{j=1}^{p} \beta_j g_j(x_i) + \beta_0
\end{equation}
The gene weights $\boldsymbol{\beta}$ are estimated by minimizing the \textcolor{black}{elastic net} cost:
\begin{equation}
    \hat{\boldsymbol{\beta}} = \arg\min_{\boldsymbol{\beta}} \left\{ \frac{1}{n} \sum_{i=1}^{n} \left( y_i - \hat{y}_i \right)^2 + \lambda_1 \|\boldsymbol{\beta}\|_1 + \lambda_2 \|\boldsymbol{\beta}\|_2^2 \right\}
\end{equation}
where $\lambda_1$ and $\lambda_2$ control the L1 and L2 penalty contributions. This procedure is applied independently within each population. Isolated fitness is defined by the \textcolor{black}{root mean squared error (RMSE)} over a 5-fold cross-validation scheme to prevent premature convergence to brittle symbolic forms.

\subsection{Ensemble Fitness through Multi-Population Symbolic Fusion}

To capture the interactions among structurally separated populations, DIGSP constructs an ensemble model that aggregates predictions from top-performing individuals in each population. A second ENR is used to learn the optimal weights for fusing these inter-population predictions. This symbolic ensemble preserves the physical separability of component mechanisms while enabling global expressivity through symbolic superposition.

\subsection{Adaptive Hierarchical Symbolic Abstraction Mechanism (AHSAM)}

AHSAM is invoked when no improvement is observed for $T$ generations in all populations. It enables hierarchical recomposition of symbolic components through statistical abstraction and selective injection. The mechanism consists of two phases:

\subsubsection*{Phase I: ANOVA-Based Symbolic Filtering}

After $T$ generation of no improvement in overall fitness, the outputs of all individuals are evaluated across the training set and subjected to one-way ANOVA to determine whether any individual contributes statistically distinct behavior. The total variance is decomposed as:
\begin{equation}
SS_{\text{total}} = SS_{\text{between}} + SS_{\text{within}}
\end{equation}
where $SS_{\text{between}}$ captures the variance of mean predictions across individuals and $SS_{\text{within}}$ quantifies variance within each individual’s outputs. The F-statistic is computed as:
\begin{equation}
    F = \frac{SS_{\text{between}} / (k-1)}{SS_{\text{within}} / (n - k)}
\end{equation}
where $k$ is the number of individuals and $n$ the number of data samples. Individuals with $p$-values $\leq 0.05$ are retained as candidates for abstraction.

\subsubsection*{Phase II: Symbolic Compression and Cross-Population Injection}

Each retained individual $g^{(i)}(x)$ is abstracted into a symbolic construct $z_i = g^{(i)}(x)$ and injected into the terminal set of all populations. Before injection, DIGSP evaluates the standalone utility of $z_i$ by computing its RMSE and comparing it to the average RMSE of the current population. If $z_i$ does not degrade predictive accuracy, it is retained. If not, a structural pruning step removes destabilizing subtrees until $z_i$ either improves or matches the baseline.

This abstraction-reinjection cycle encodes symbolic superposition by permitting learned expressions from one structural group to inform the symbolic evolution of others. The resulting feature space becomes enriched with cross-mechanism constructs while preserving hierarchical interpretability.

\subsection{Termination Criteria}

The evolutionary process terminates if either: (i) the maximum number of generations is reached, or (ii) no improvement occurs for a fixed number of generations following one or more AHSAM activations. The final model is selected based on ensemble RMSE performance over the validation set.

\section{Results and Discussion} \label{Section:Results_Discussion}

This section evaluates the performance of DIGSP in modeling the nonlinear shear strength of SFRC beams and compares it to a BGP model that evolves all symbolic genes within a single undivided population. Both models are executed across 30 independent runs, with each run involving a randomized 65\% training, 10\% validation, and 25\% test partition. To ensure comparability, the total number of genes per individual is kept constant, and both models employed identical functional operators, crossover and mutation rates, and maximum generation budgets.

\begin{table}[!ht]
\centering
\caption{Parameter configuration for BGP and DIGP models.}
\label{tab:parameters}
\small
{%
\color{black}%
\arrayrulecolor{black}%
\begin{tabular}{lcc}
\toprule
\textbf{Parameter} & \textbf{BGP} & \textbf{DIGP} \\
\midrule
Number of Runs           & 30    & 30\\
Number of Populations    & 1     & 3    \\
Population Size          & 50    & 50       \\
Max Generations          & 300   & 300     \\
Stall Generation         & 30    & 30    \\
AHSAM Trigger Generation  & - & 25 \\
Genes per Individual     & 9    & 3   \\
Max Tree Depth           & 15    & 15  \\
Initialization           & Half-and-Half & Half-and-Half  \\
$p_{c}$            & 0.84    & 0.84  \\
$p_{m}$             & 0.14    & 0.14  \\
$p_{r}$         & 0.02     & 0.02    \\
Constant Range                        & [-10, 10] & [-10, 10]  \\
Functional nodes		& $+$ $-$ $/$ $\times$ & $+$ $-$ $/$ $\times$  \\
\bottomrule
\end{tabular}
}%
\end{table}

Table~\ref{tab:parameters} summarizes the key hyperparameters configured for the BGP and DIGSP models. Both algorithms are run for a maximum of 300 generations with early stopping triggered by 30 consecutive stagnant generations. The primary architectural distinction lies in DIGSP’s use of three co-evolving populations, each encoding three symbolic genes per individual, in contrast to BGP’s single population with nine genes per individual. This multi-population configuration enables DIGSP to isolate material-specific feature subsets, facilitating the evolution of specialized substructures that are symbolically abstracted via the AHSAM when stagnation occurs.

The AHSAM trigger is set to 25 generations in DIGSP runs, ensuring that symbolic abstraction is not applied prematurely. The rest of the evolutionary parameters, such as crossover ($p_{c} = 0.84$), mutation ($p_{m} = 0.14$), and reproduction ($p_{r} = 0.02$), are aligned across both models to ensure comparability. The function set included the four basic arithmetic operations $\{+, -, \times, \div\}$, with numerical constants sampled uniformly from the range $[-10, 10]$. Both models used the half-and-half initialization method and a maximum tree depth of 15 to prevent overly complex symbolic structures early in evolution.

\begin{figure*}[ht]
\centering
\subfloat[Train Fitness]{\includegraphics[width=0.32\textwidth]{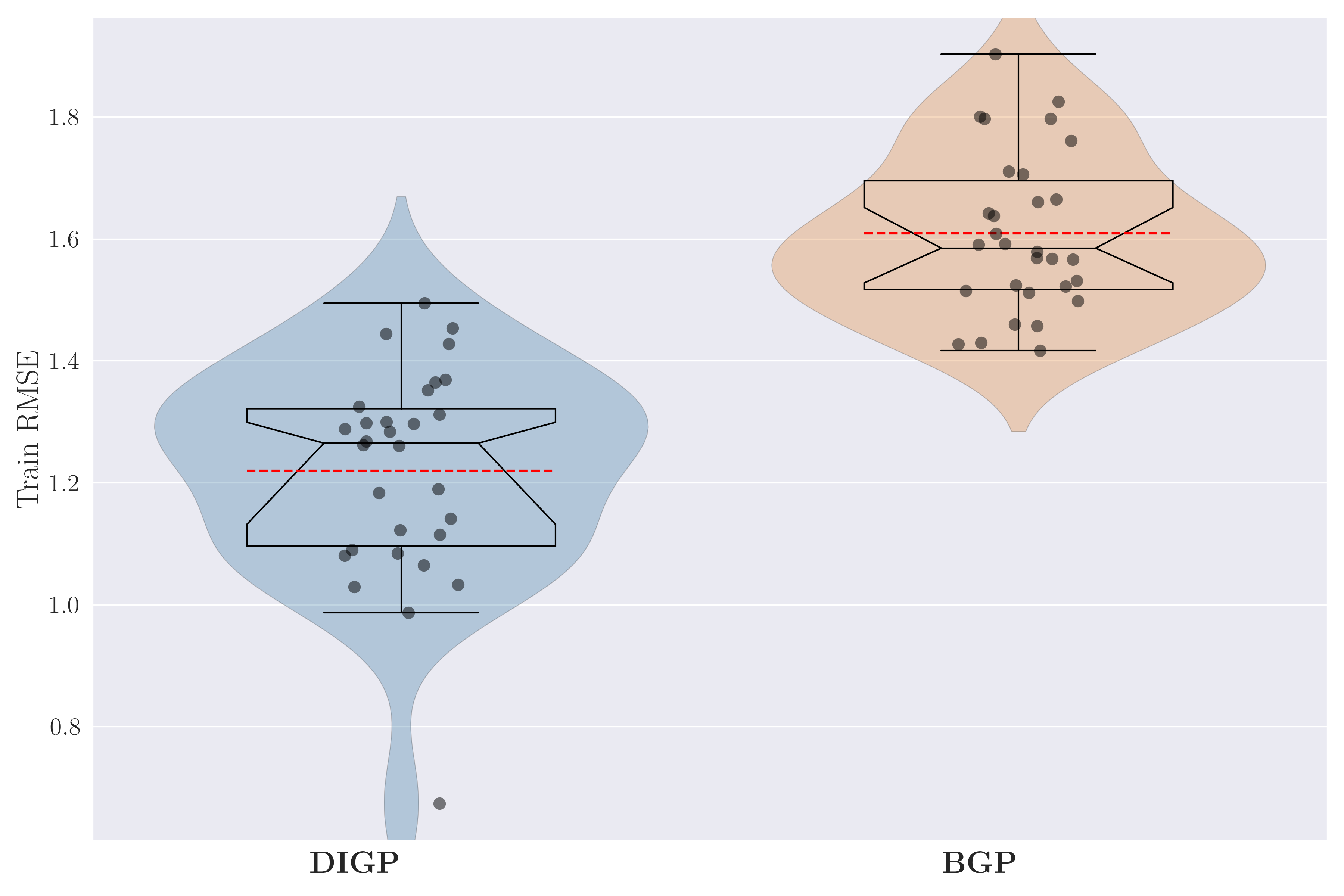}\label{fig:train_fitness}}%
\hfill
\subfloat[Validation Fitness]{\includegraphics[width=0.32\textwidth]{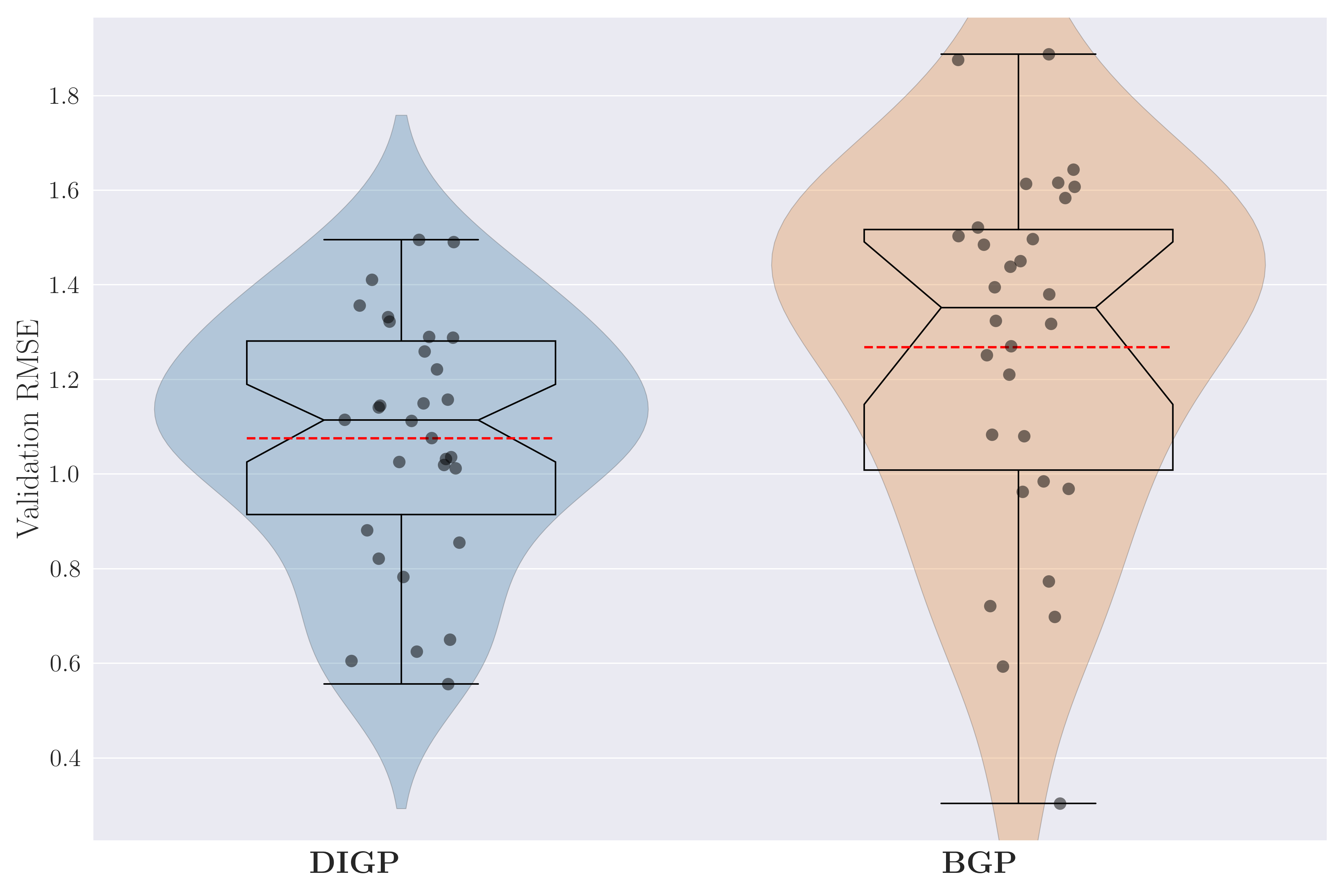}\label{fig:validation_fitness}}%
\hfill
\subfloat[Test Fitness]{\includegraphics[width=0.32\textwidth]{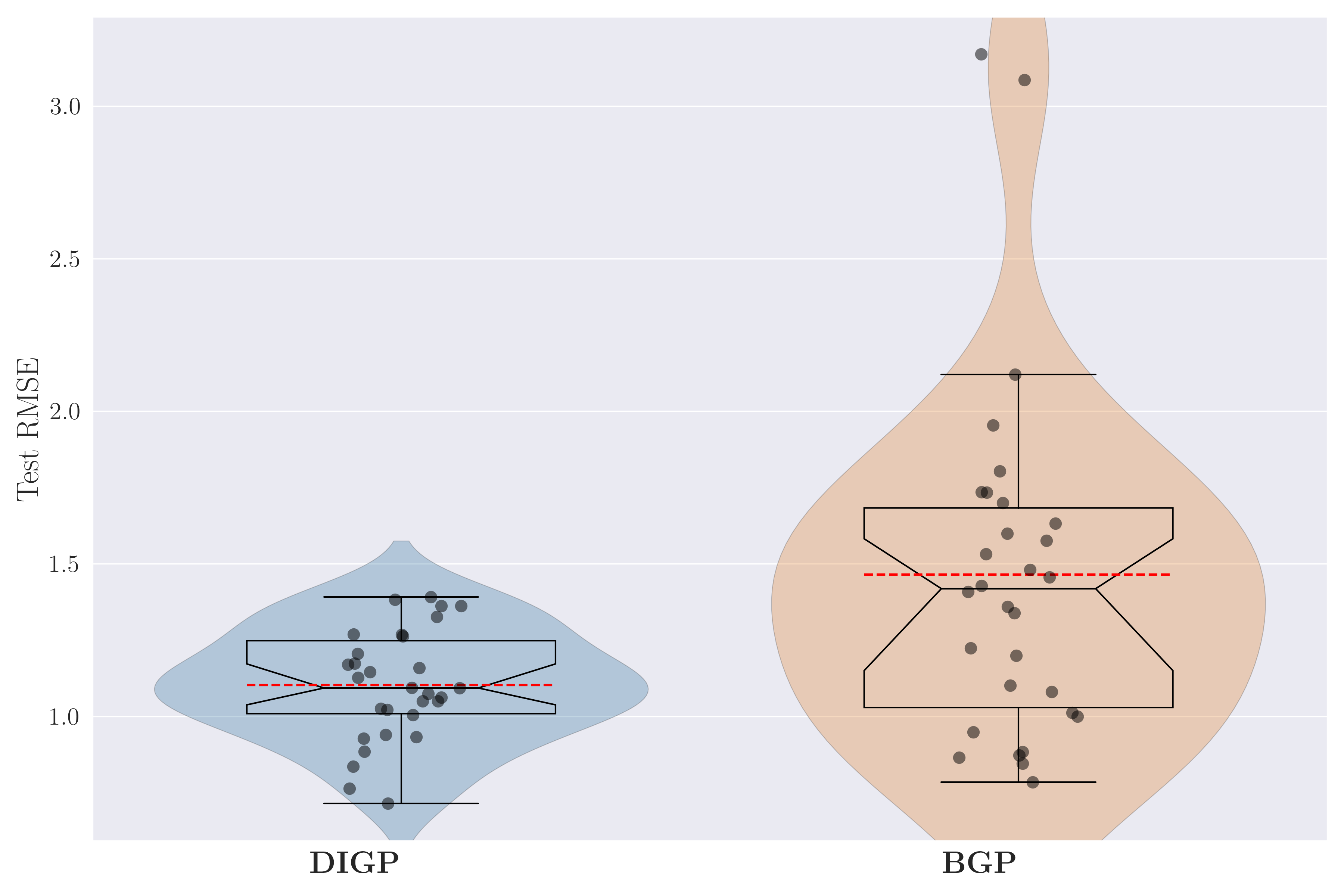}\label{fig:test_fitness}}

\caption[Raincloud plots comparing DIGSP and BGP performance]{Raincloud plots comparing the fitness distributions of \textcolor{black}{DIGSP} and baseline \textcolor{black}{BGP} over 30 runs. Subfigures show the distribution of (a) training RMSE, (b) validation RMSE, and (c) test RMSE.}
\label{fig:fitness_comparison}
\end{figure*}

Figure~\ref{fig:fitness_comparison} presents raincloud plots comparing the training, validation, and test RMSE distributions of BGP and DIGP. \textcolor{black}{The visualization employs violin plots, boxplots, and jittered scatter points, where each subplot shows the kernel density estimate of the distribution (violin), the interquartile range and median (boxplot), and individual model instances across 30 runs (scatter).} The left panel (\ref{fig:train_fitness}) shows that DIGSP achieved noticeably lower training RMSE compared to BGP, with the mean and median located below 1.25, in contrast to BGP’s values clustering around 1.6. This suggests that the evolutionary dynamics in DIGSP, facilitated by isolated gene partitions and periodic symbolic abstraction, enhanced its ability to capture structural regularities in the training data.

The Wilcoxon rank-sum test on training RMSEs further confirmed the statistical superiority of DIGSP, yielding a $p$-value of 0.007, thereby rejecting the null hypothesis at the 0.05 significance level. This indicates that DIGSP consistently learns more expressive or better-generalizing symbolic expressions from the training data, despite being constrained to smaller gene structures per individual.

The central panel of Figure~\ref{fig:fitness_comparison} (\ref{fig:validation_fitness}) depicts the validation RMSE distribution. While DIGSP still shows a central tendency towards lower error compared to BGP, the gap is less pronounced. The Wilcoxon rank-sum test resulted in a $p$-value of 0.23, implying that the null hypothesis cannot be rejected with high confidence. This outcome is likely attributable to the small size of the validation set (10\% of total data), which introduces high variance in the validation estimates and reduces statistical power. Given that evolutionary selection in both models partially relies on validation fitness to prevent overfitting, this finding highlights the limitations of small validation sets in reliably differentiating between competing symbolic structures.

\begin{table}[h]
\centering
\caption{\textcolor{black}{Overall performance across 30 runs, including median [IQR] and mean $\pm$ SD of obtained RMSEs. }}
\label{tab:results_summary}
\small
{%
\color{black}%
\arrayrulecolor{black}%
\begin{tabular}{l l c c}
\toprule
\textbf{Metric} & \textbf{Model} & \textbf{Median [IQR]} & \textbf{Mean $\pm$ SD} \\
\midrule
Train & DIGSP & 1.265 [1.096--1.321] & 1.220 $\pm$ 0.173 \\
Train & BGP   & 1.585 [1.517--1.695] & 1.609 $\pm$ 0.131 \\
Validation & DIGSP & 1.114 [0.914--1.281] & 1.075 $\pm$ 0.260 \\
Validation & BGP   & 1.352 [1.008--1.516] & 1.268 $\pm$ 0.383 \\
Test & DIGSP & 1.093 [1.009--1.248] & 1.103 $\pm$ 0.181 \\
Test & BGP   & 1.419 [1.029--1.683] & 1.464 $\pm$ 0.575 \\
\bottomrule
\end{tabular}
}
\end{table}

The right panel of Figure~\ref{fig:fitness_comparison} (\ref{fig:test_fitness}) provides the most critical insight into model generalization. Here, DIGSP again outperformed BGP, achieving lower test RMSE values across nearly all runs. The dispersion of DIGSP’s distribution is notably narrower, with reduced presence of outliers. This reinforces the hypothesis that multi-population symbolic partitioning and abstraction via AHSAM reduces overfitting while retaining expressive power.

\begin{table*}[h]
\centering
\caption{\textcolor{black}{Model parsimony and computational profile across 30 independent runs. Tree size is reported as node count (operators$+$variables$+$constants) extracted from the closed-form expressions.}}
\label{tab:model_parsimony}
\small
{%
\color{black}%
\arrayrulecolor{black}%
\begin{tabular}{lccccccccc}
\toprule
\textbf{Model} & \textbf{\# Terms} & \textbf{Tree Size (nodes)} & \textbf{Operator Count} & \textbf{\shortstack{Train Time / Run\\(sec) Median [IQR]}} & \textbf{\shortstack{Train Time / Run\\(sec) Mean $\pm$ SD}} & \textbf{\shortstack{Total\\(30 runs) (h)}} \\
\midrule
DIGSP & 7 [6--8] &  41 [35--68] & 138 [120--175] & 308.90 [235.25--397.17] & 320.43 $\pm$ 52.82 & 3.23 \\
BGP   & 21 [19--23]    &  204 [177--259]   & 26 [21--46]   & 238.34 [222.32--380.92] & 301.61 $\pm$ 120.35 & 2.51 \\
\bottomrule
\end{tabular}
}%
\end{table*}

\begin{figure*}[t]
\centering
% -------- Row 1 --------
\subfloat[\small \textcolor{black}{DIGSP (mean)}]{%
  \includegraphics[width=0.5\textwidth]{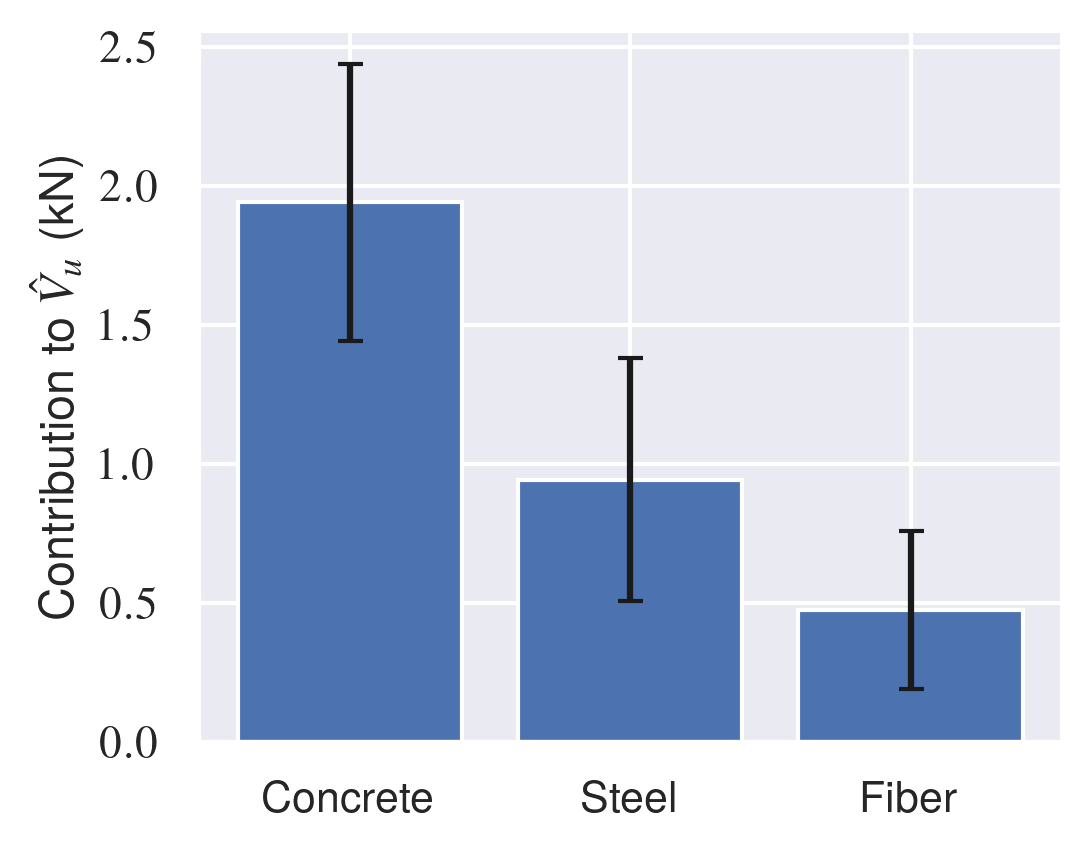}%
  \label{fig:cont_digsp_mean}}
\hfill
\subfloat[\small \textcolor{black}{BGP (mean)}]{%
  \includegraphics[width=0.5\textwidth]{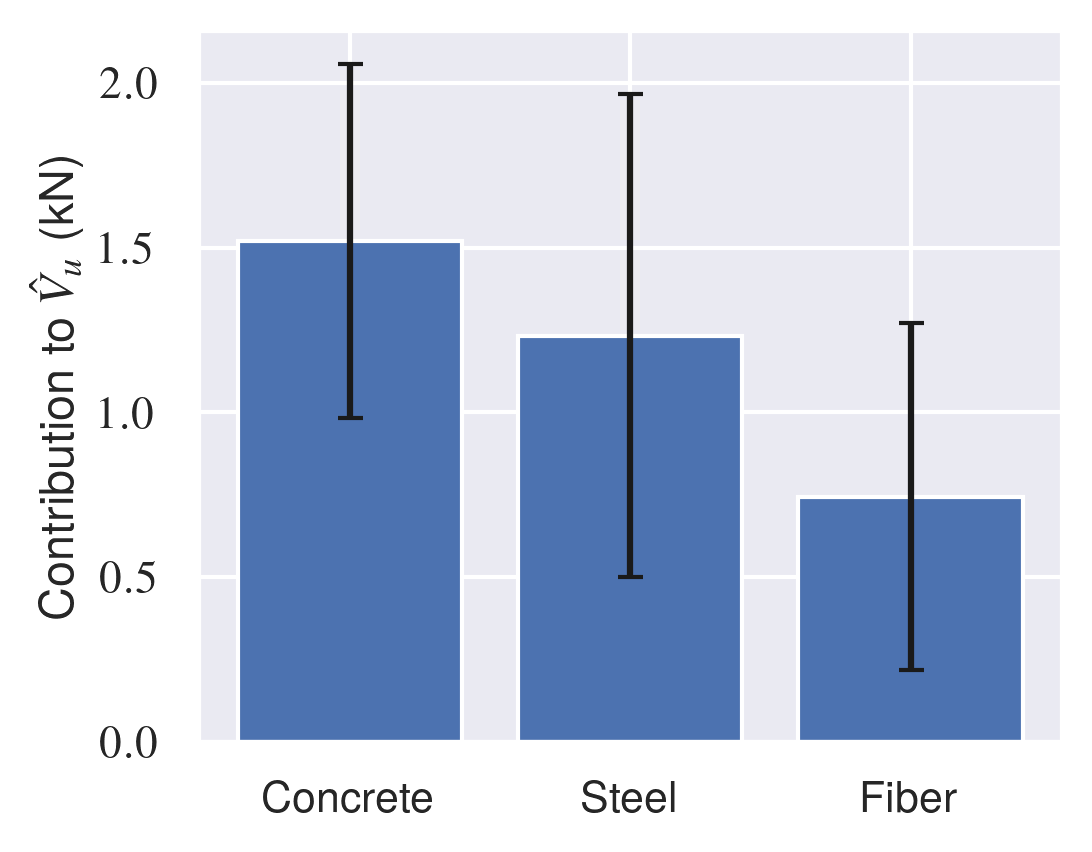}%
  \label{fig:cont_bgp_mean}}

\vspace{0.7em}

% -------- Row 2 --------
\subfloat[\small \textcolor{black}{DIGSP (runs)}]{%
  \includegraphics[width=0.5\textwidth]{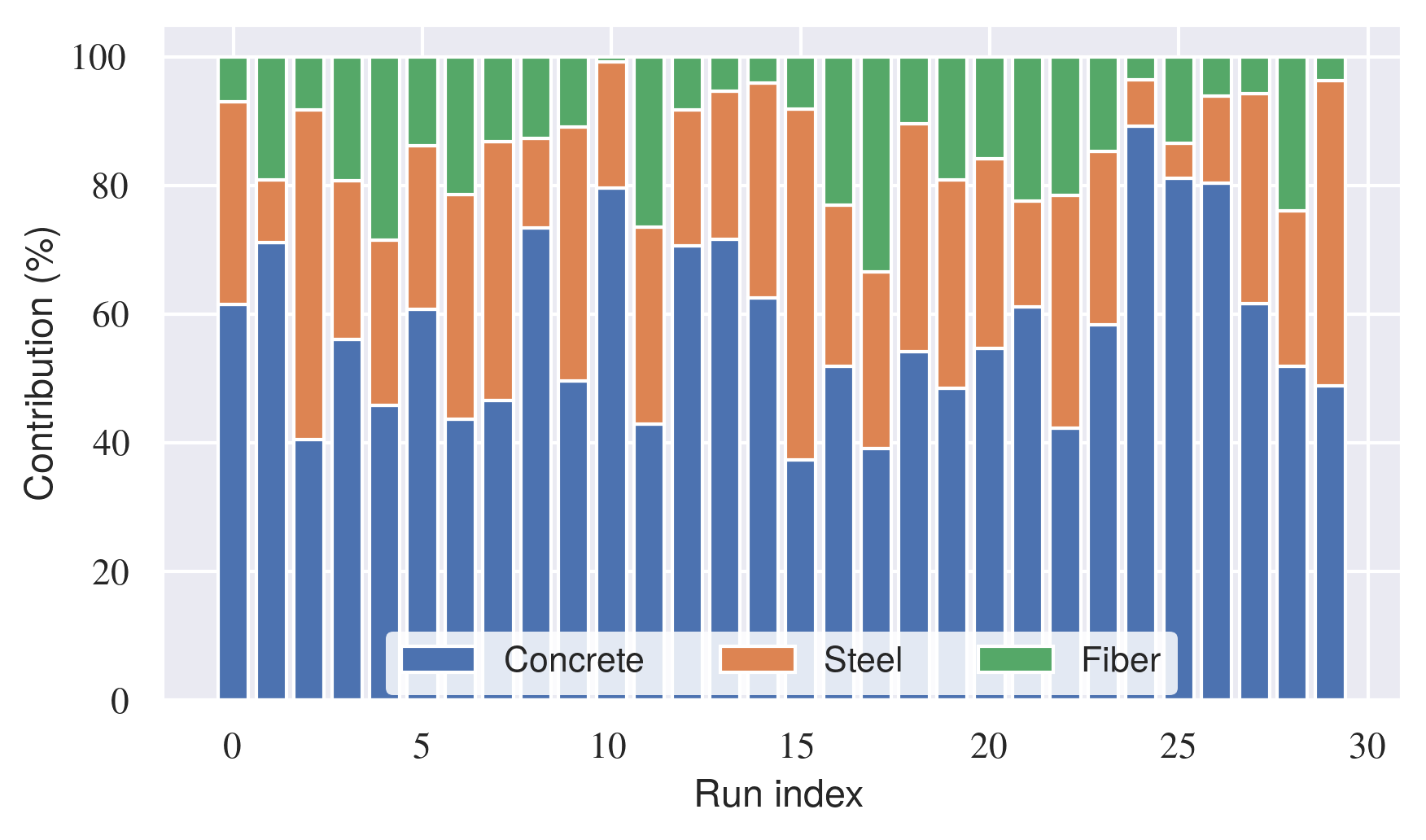}%
  \label{fig:cont_digsp_runs}}
\hfill
\subfloat[\small \textcolor{black}{BGP (runs)}]{%
  \includegraphics[width=0.5\textwidth]{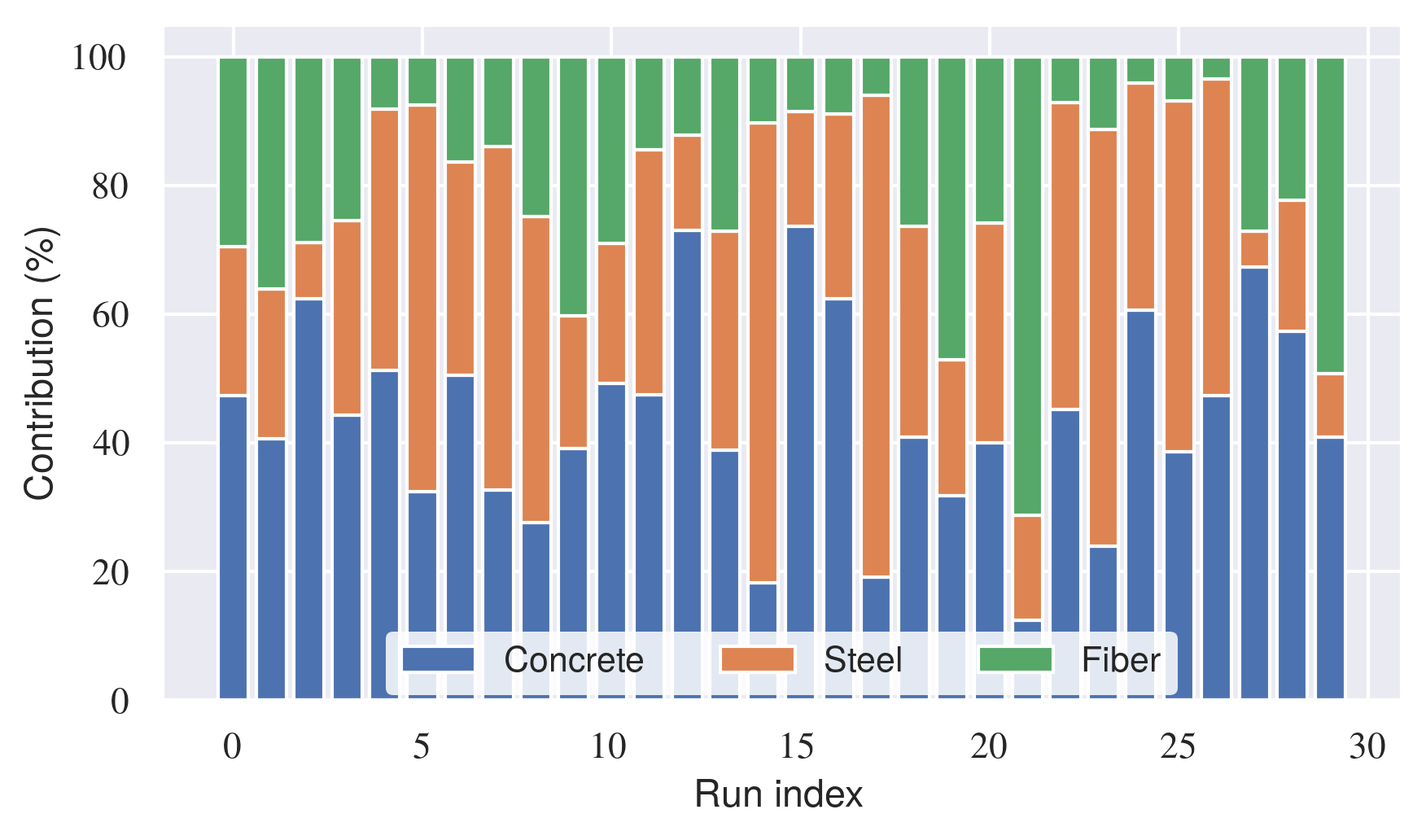}%
  \label{fig:cont_bgp_runs}}

\caption{\textcolor{black}{Mechanism-level contributions to the predicted ultimate shear strength, $\hat V_u$, evaluated at the dataset-median input vector. Components correspond to concrete, steel, and fiber. Panels (a)–(b) report mean contributions (kN) across 30 independent runs with $\pm$1 standard-deviation error bars. Panels (c)–(d) show the per-run percentage breakdown (\%) to reveal run-to-run variability.}}
\label{fig_contribution}
\end{figure*}

\textcolor{black}{Across 30 independent runs, DIGSP consistently achieved lower RMSE than the BGP on train, validation, and test splits (Table~\ref{tab:results_summary}). On the test split, the median and interquartile range (IQR) RMSE for DIGSP is 1.093 [1.009--1.248] versus 1.419 [1.029--1.683] for BGP, with corresponding means $\pm$ standard deviation (SD) of 1.103 $\pm$ 0.181 and 1.464 $\pm$ 0.575, respectively. Similar gaps are observed on validation (DIGSP: 1.114 [0.914--1.281]; BGP: 1.352 [1.008--1.516]) and train (DIGSP: 1.265 [1.096--1.321]; BGP: 1.585 [1.517--1.695]) splits, indicating that the improvement is not confined to a particular partition. Lower central tendency and reduced dispersion for DIGSP suggested more stable generalization across runs.}

\textcolor{black}{To quantify superiority on held-out data, a paired analysis is performed on the per-run test RMSE using the difference $\Delta=\mathrm{RMSE}_{\mathrm{BGP}}-\mathrm{RMSE}_{\mathrm{DIGSP}}$ (positive values favor DIGSP). The distribution of $\Delta$ had a median [IQR] of 0.213 [0.020--0.707] and a mean $\pm$ SD of 0.362 $\pm$ 0.551, with a bootstrap 95\% confidence interval (CI) for the mean of [0.169,\,0.573]. DIGSP outperformed BGP in 76.7\% of runs (23/30). A one-sided Wilcoxon signed-rank test (alternative: median $\Delta>0$) yielded $W=379$ and $p=9.32\times10^{-4}$, with a rank-biserial effect size $r=0.533$, indicating a statistically significant and practically moderate improvement.}

\textcolor{black}{The parsimony analysis in Table~\ref{tab:model_parsimony} shows that DIGSP produced markedly more compact closed-form models than the BGP across 30 independent runs. In particular, the median number of top-level terms is lower for DIGSP than for BGP, and the median tree size (node count) is substantially smaller (e.g., 41 versus 204 nodes), indicating that the domain-informed partitioning favored concise superpositional expressions that are easier to interpret.}

\textcolor{black}{From a computational standpoint, the training-time profile remained practical. While DIGSP incurred a modest per-run overhead relative to BGP (median $\sim$309\,s versus $\sim$238\,s and mean $320\pm53$\,s versus $302\pm120$\,s), the total 30-run wall-clock remained within a few hours (3.23\,h for DIGSP vs.\ 2.51\,h for BGP). Overall, the increase in compute is limited compared with the observed gains in parsimony and the generalization improvements reported earlier, supporting the practicality of the physics-informed search strategy.}

\textcolor{black}{Figure~\ref{fig_contribution} provides a mechanism‐level (concrete, steel, and fiber) attribution of the predicted shear capacity, $\hat V_u$, evaluated at the dataset–median input vector. In the mean summaries (panels~\ref{fig:cont_digsp_mean}–\ref{fig:cont_bgp_mean}), DIGSP yields a clear and physically consistent ranking of contributions—concrete $>$ steel $>$ fiber—with markedly tighter uncertainty bands across 30 independent runs that indicate superior parameter identifiability and robustness of the superpositional decomposition. The per–run percentage stacks (panels~\ref{fig:cont_digsp_runs}–\ref{fig:cont_bgp_runs}) further show that DIGSP concentrates mass on the concrete mechanism with comparatively modest run‐to‐run drift in the steel and fiber shares, whereas BGP exhibits substantially larger dispersion and occasional ordering reversals. These patterns align with the expected mechanics of SFRC beam shear—matrix (concrete) action dominating, reinforcement providing secondary load paths, and fibers contributing post‐cracking bridging—and demonstrate that the domain‐informed partitioning in DIGSP stabilizes the search, preserves additivity of effects, and delivers interpretable mechanism‐level diagnostics that a monolithic GP baseline cannot reliably recover.}

\begin{table*}[t]
\centering
\caption{\textcolor{black}{One-point elasticities $S_i$ of $\hat V_u$ at the dataset-median input, using a 1\% forward perturbation ($\epsilon=0.01$) on the closed-form expressions. Values are dimensionless and summarized across 30 runs by median [IQR] and mean~$\pm$~SD. Positive elasticities increase $\hat V_u$; negative decrease it.}}
\label{tab:one_point_sens}
\small
{\color{black}
\arrayrulecolor{black}
\begin{tabular}{l c c c c c}
\toprule
\multirow{2}{*}{\textbf{Variable} $x_i$} & \multirow{2}{*}{\textbf{Expected sign}} &
\multicolumn{2}{c}{\textbf{DIGSP} ($N{=}30$)} & \multicolumn{2}{c}{\textbf{BGP} ($N{=}30$)} \\
\cmidrule(lr){3-4}\cmidrule(lr){5-6}
& & \textbf{Median [IQR]} & \textbf{Mean $\pm$ SD} & \textbf{Median [IQR]} & \textbf{Mean $\pm$ SD} \\
\midrule
$V_f$ (fiber vol.\ \%)   & $+$ & $0.12\;[0.09,\,0.16]$ & $0.12 \pm 0.04$ & $0.08\;[0.02,\,0.14]$ & $0.09 \pm 0.08$ \\
$f_c'$ (concrete strength, MPa) & $+$ & $0.33\;[0.27,\,0.39]$ & $0.33 \pm 0.07$ & $0.24\;[0.08,\,0.37]$ & $0.23 \pm 0.14$ \\
$\rho$ (long.\ steel ratio)     & $+$ & $0.21\;[0.16,\,0.26]$ & $0.21 \pm 0.06$ & $0.14\;[0.04,\,0.24]$ & $0.15 \pm 0.13$ \\
$a/d$ (shear span / depth)      & $-$ & $-0.28\;[-0.34,\,-0.22]$ & $-0.28 \pm 0.07$ & $-0.18\;[-0.31,\,-0.05]$ & $-0.19 \pm 0.15$ \\
\bottomrule
\end{tabular}
}
\end{table*}

\textcolor{black}{To quantify local, mechanism-consistent sensitivities, one-point elasticities of the predicted shear capacity are computed at the dataset-median operating point $\tilde{\boldsymbol{x}}$ for four physically salient inputs: fiber volume fraction $V_f$, concrete strength $f'_c$, longitudinal reinforcement ratio $\rho$, and shear-span ratio $a/d$. For each run and model, the elasticity of $x_i$ is obtained by a forward finite difference on the closed-form expression (no retraining):}

\textcolor{black}{
\begin{equation}
S_i(\tilde{\boldsymbol{x}})\;\approx\;
\frac{\hat V_u\!\big(\tilde{x}_1,\ldots,\tilde{x}_i(1+\epsilon),\ldots\big)-\hat V_u(\tilde{\boldsymbol{x}})}{\epsilon\,\hat V_u(\tilde{\boldsymbol{x}})}\,.
\end{equation}
}

\textcolor{black}{where, $S_i$ denotes the dimensionless elasticity of variable $x_i$; $\hat V_u(\cdot)$ is the predicted ultimate shear strength; $\tilde{\boldsymbol{x}}$ is the dataset-median input vector; $\tilde{x}_i$ is its $i$th component; and $\epsilon$ is the relative perturbation. In the limit $\epsilon\to 0$, $S_i(\tilde{\boldsymbol{x}})\to \big(\partial \hat V_u/\partial x_i\big),(\tilde{x}_i/\hat V_u(\tilde{\boldsymbol{x}}))$. We summarized results over 30 runs by the median [IQR] and mean~$\pm$~SD in Table~\ref{tab:one_point_sens}.}

\textcolor{black}{Across runs, DIGSP produced a stable and physically coherent ordering: $f'_c$ exhibited the largest positive elasticity, $\rho$ and $V_f$ are positive but smaller, and $a/d$ is consistently negative (Table~\ref{tab:one_point_sens}). For illustration, the DIGSP median elasticity for $V_f$ is about $0.12$ (versus $\approx 0.08$ for BGP), and $a/d$ is more strongly negative for DIGSP (median $\approx -0.28$) than for BGP (median $\approx -0.18$). The tighter IQRs/SDs under DIGSP indicated improved parameter identifiability and a more regular local response surface around $\tilde{\boldsymbol{x}}$, consistent with the intended superpositional, domain-informed structure.}

These results validate the efficacy of DIGSP in modeling complex regression relationships. Its architecture exploits structural redundancy across feature subsets and introduces semantic compression via AHSAM only when evolutionary progress stalls, thereby improving both efficiency and generalization. The experimental design—aligned in terms of population size, operator probabilities, functional nodes, and training budget—ensures that the observed performance gains stem from DIGSP’s algorithmic innovations rather than from parameter tuning or increased model complexity. The overall findings position DIGSP as a more robust alternative to conventional symbolic regression models, especially in domains where variable interactions are intricate and overfitting is prevalent.

\section{Conclusion} \label{Section:Conclusion}

In this study \textcolor{black}{the} DIGSP framework \textcolor{black}{is introduced}, a SR framework designed to model physical systems that exhibit separable structural mechanisms. DIGSP operationalizes the principle of superposition by evolving domain-partitioned populations of symbolic expressions and hierarchically recomposing them through an adaptive abstraction mechanism. Applied to the problem of predicting the shear strength of SFRC beams, DIGSP demonstrated substantial advantages in convergence stability and test generalization compared to a BGP model.

By structuring the search space according to physically meaningful feature groupings—concrete, steel, and fiber mechanics—DIGSP enables each subpopulation to specialize over a well-defined subset of the input domain. The use of ENR ensures robust gene fusion within individuals and across populations, while AHSAM selectively transfers statistically validated sub-expressions into all populations only upon global stagnation. This abstraction-reinjection cycle allows DIGSP to build complex yet interpretable symbolic models in a bottom-up fashion, reflecting the physical compositionality of SFRC shear behavior.

Empirical results across 30 independent runs confirmed the statistical superiority of DIGSP over BGP in both training and test RMSE, with the latter achieving significantly lower error dispersion and higher stability. These improvements are obtained without altering the underlying operators, depth limits, or training budgets—highlighting that the performance gain stems from the architectural principles of domain-informed partitioning and symbolic superposition.

The DIGSP framework provides a generalizable modeling strategy for structural systems in which different material or geometric components contribute independently or semi-independently to global behavior. Its modular evolution, statistical abstraction, and compositional symbolic structure offer an interpretable and efficient alternative to black-box regressors in engineering contexts where physical consistency and transparency are essential. Future work will investigate extending DIGSP to spatiotemporal domains, incorporating physics-based constraints into the abstraction cycle, and deploying the method in other fields where superpositional modeling is appropriate.

\bibliography{Refs}

%
%% \appendix

\end{document}